\documentclass[letterpaper, 10 pt, conference]{ieeeconf}

\IEEEoverridecommandlockouts
\usepackage{amsmath}
\usepackage{amssymb}
\usepackage{graphicx}
\usepackage{booktabs}
\usepackage{multirow}
\usepackage{array}
\usepackage{pifont}
\usepackage{bm}

\usepackage{array}
\usepackage{multirow}

\newcommand{\cmark}{\ding{51}}
\newcommand{\xmark}{\ding{55}}

\title{\LARGE \bf
Video2DoorTraversal: Push Door Traversal via Simulated Door Twins
}

\author{
Xincheng Tang$^{1}$, Yiji Chen$^{1}$, Youhan Xie$^{1}$, Wanyu Li$^{1}$, Zhengjie Shu$^{1}$,\\
Lai Jiang$^{1}$, Wenkang Hu$^{1}$, Yitong Li$^{1}$, Jinchuang Zhang$^{3}$,
Xibin Song$^{2}$, and Ruigang Yang$^{1,3\dagger}$%
\thanks{$^{1}$Shanghai Jiao Tong University.
E-mail: \{tangxincheng, ryang2\}@sjtu.edu.cn}%
\thanks{$^{2}$Shandong University.}%
\thanks{$^{3}$NeoWa Robotics.}%
\thanks{$^{\dagger}$Corresponding author.}%
}

\begin{document}

\maketitle
\thispagestyle{empty}
\pagestyle{empty}

\begin{abstract}
Door opening and traversal is a long-horizon loco-manipulation task that requires precise handle interaction and coordinated base-arm control. We present \textbf{Video2DoorTraversal}, a single-video real-to-sim-to-real framework for wheel-legged mobile manipulators. Given one RGB video of a real door, \textbf{DoorTwin} reconstructs an instance-aligned, articulated, and simulation-ready door twin with realistic geometry and appearance. A simulation-in-the-loop agent converts the recovered articulation into a parameterized skill program and iteratively refines failed rollouts to generate physically executable demonstrations. These demonstrations are used to train \textbf{ArticuACT}, a dual-depth policy that predicts coordinated base, arm, and gripper commands using robot-centric camera conditioning and interaction-aware supervision. With all perception and policy inference running onboard, the system achieves a \(96.57\%\) average success rate across five real doors and an \(80.95\%\) zero-shot success rate on structurally similar unseen doors, while completing the full approach, opening, and traversal sequence in approximately \(13\,\mathrm{s}\) on average.

Project Page: https://video2doortraversal.github.io/.
\end{abstract}

\section{Introduction}

Autonomous door opening and traversal is essential for mobile robots operating in human-centered environments. Unlike isolated handle manipulation or door pushing, a legged or wheel-legged manipulator must approach the door, unlock the handle, coordinate its base and arm while the panel moves, and pass through a narrow doorway without collision. The task is therefore a long-horizon, contact-rich loco-manipulation problem that tightly couples mobility and manipulation~\cite{zhang2024leggeddoor,sleiman2024guidedrl,wang2025doorbot,xue2025doorman}.

Existing methods solve only parts of this problem. DoorGym~\cite{urakami2019doorgym}, UniDoorManip~\cite{li2024unidoormanip}, and ArticuBot~\cite{wang2025articubot} learn articulated-object manipulation from large-scale simulation, but mainly focus on opening rather than doorway traversal. Adaptive mobile manipulation~\cite{xiong2024adaptive} and DoorBot~\cite{wang2025doorbot} improve real-world robustness through online adaptation or haptic feedback, while legged and humanoid systems achieve complete traversal through reinforcement learning or teacher--student training~\cite{zhang2024leggeddoor,sleiman2024guidedrl,xue2025doorman}. However, these approaches typically rely on preconstructed or procedurally generated doors, manually specified task structure, extensive reward design, or additional real-world adaptation, rather than constructing a task-specific simulator from the observed door.

Real-to-sim-to-real methods reduce robot data collection by transferring task information from human videos or real observations into simulation. HUMAN2SIM2ROBOT~\cite{lum2025human2sim2robot}, X-SIM~\cite{dan2025xsim}, and Video2Sim2Real~\cite{han2026video2sim2real} use human demonstrations or object motion to guide embodiment-specific policy learning, but mainly target fixed-base rigid-object manipulation and require additional scene or object scans. DemoGen~\cite{xue2025demogen} generates synthetic visuomotor demonstrations, while SIMPACT~\cite{liu2026simpact} constructs a simulator from a single RGB observation for test-time action refinement. Nevertheless, establishing a practical real-to-sim-to-real pipeline for complete door traversal remains nontrivial. The central challenge is to bridge instance-specific articulation and closed-loop execution: the reconstructed door must be metrically and kinematically accurate enough to support contact-rich simulation, while the learned behavior must remain robust to perception and dynamics discrepancies during real-world deployment. We therefore formulate door traversal as an articulation-conditioned policy-learning problem, in which the recovered door twin provides a common task representation for scene reconstruction, expert generation, and policy execution.

\begin{figure}[t]
    \centering
    \includegraphics[width=\linewidth]{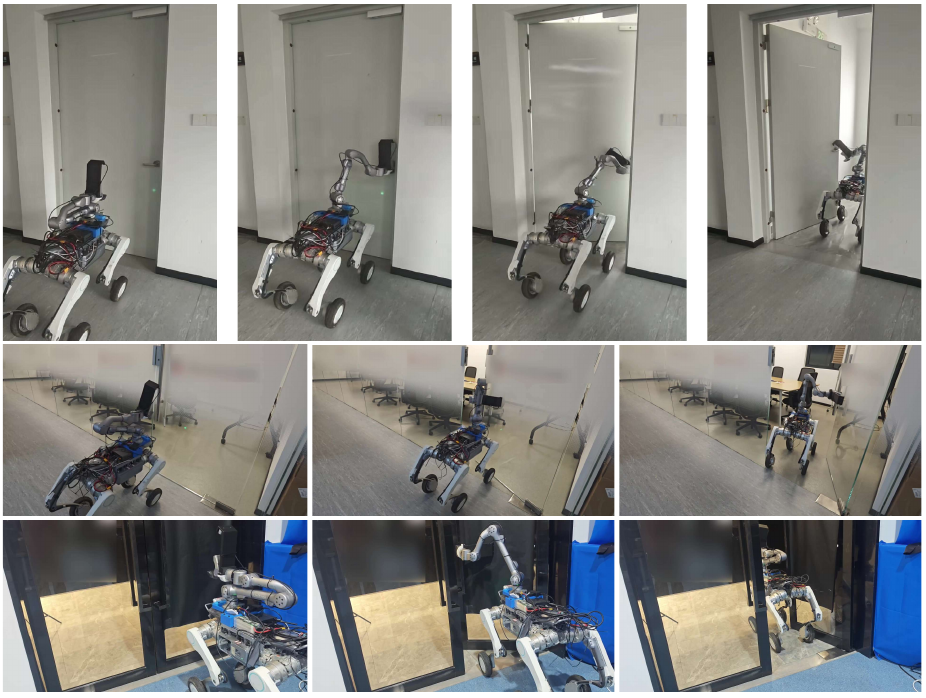}
    \caption{From a single RGB video, Video2DoorTraversal enables real-world wheel-legged door opening and traversal.}
    \label{fig:teasor}
    \vspace{-0.6cm}
\end{figure}

\begin{table*}[t]
\vspace{2mm}
\centering
\caption{System-level comparison with representative real-to-sim-to-real manipulation and door-traversal methods.}
\label{tab:system_comparison}
\setlength{\tabcolsep}{0pt}
\renewcommand{\arraystretch}{1.02}
\scriptsize
\begin{tabular*}{\textwidth}{@{\extracolsep{\fill}}llcccclc@{}}
\toprule
\textbf{Method} & \textbf{Real-to-Sim Input} & \textbf{Mobile Robot} &
\textbf{Articulated Manipulation} & \textbf{Agentic Expert} &
\textbf{Sim-to-Real} & \textbf{Perception Input} & \textbf{Door Traversal} \\
\midrule
Human2Sim2Robot & \cmark\ Human RGB video + scans & \xmark & \xmark & \xmark & \cmark & 6D object pose & \xmark \\
X-SIM            & \cmark\ Human RGB video + scans & \xmark & \xmark & \xmark & \cmark & RGB & \xmark \\
SIMPACT          & \cmark\ Single RGB image   & \xmark & \xmark & \cmark & \cmark & Initial RGB & \xmark \\
UniDoorManip     & \xmark & \cmark & \cmark & \xmark & \cmark & Point cloud & \xmark \\
Teacher-student  & \xmark & \cmark & \cmark & \xmark & \cmark & Given handle position & \cmark \\
\textbf{Ours}    & \cmark\ \textbf{Single RGB video} & \cmark & \cmark & \cmark & \cmark & \textbf{Depth} & \cmark \\
\bottomrule
\end{tabular*}
\end{table*}

To address this challenge, we present \textbf{Video2DoorTraversal},
a single-video real-to-sim-to-real framework for wheel-legged
door traversal. \textbf{DoorTwin} reconstructs an instance-aligned,
articulated, and simulation-ready door twin from a single RGB
video. A simulation-in-the-loop agent then converts the recovered
articulation into a parameterized traversal program and iteratively
refines it through rollout diagnosis and simulator verification,
producing executable demonstrations without human teleoperation.
Finally, \textbf{ArticuACT} learns coordinated base and arm control from onboard dual-view depth observations, using
robot-centric Pl\"ucker conditioning and interaction-state
supervision to improve geometric grounding and contact-aware
execution. Unlike approaches that rely on human-motion
retargeting, online adaptation, or externally provided
door poses, our framework derives the task from a single RGB
video and executes the complete opening-to-traversal behavior
using onboard perception.

The main contributions of this work are as follows:
\begin{enumerate}
    \item We introduce a single-video real-to-sim-to-real pipeline with \textbf{DoorTwin}, which reconstructs an instance-aligned, articulated, and simulation-ready door twin from one RGB video.

    \item We propose a simulation-in-the-loop agentic expert trajectory generation method and \textbf{ArticuACT}, a dual-depth policy for closed-loop base--arm coordination with robot-centric geometric conditioning and interaction-aware supervision.

    \item We extensively validate the complete system in simulation and on real robots, achieving \(96.57\%\) average success across five real doors, and \(80.95\%\) zero-shot success across structurally similar unseen doors, with an average execution time of approximately \(13\,\mathrm{s}\).
\end{enumerate}

\section{Related Work}

\subsection{Door Opening and Traversal}

Door manipulation has been studied through simulation, online adaptation, and whole-body control. DoorGym~\cite{urakami2019doorgym}, UniDoorManip~\cite{li2024unidoormanip}, and ArticuBot~\cite{wang2025articubot} learn door or articulated-object opening from large-scale simulation, but mainly assume preconstructed assets and do not address complete doorway traversal. Real-world systems improve robustness through online adaptation or haptic feedback~\cite{xiong2024adaptive,wang2025doorbot}, while legged and humanoid platforms achieve full traversal using reinforcement learning or teacher--student training~\cite{zhang2024leggeddoor,sleiman2024guidedrl,xue2025doorman}. These methods typically rely on procedurally generated doors, external geometric inputs, task-specific rewards, or additional real-world adaptation. In contrast, our method targets the complete opening-to-traversal task and learns closed-loop base--arm coordination from onboard depth, without external geometric inputs or per-door real-world adaptation.

\subsection{Real-to-Sim-to-Real Robot Learning}

Real-to-sim-to-real methods transfer task information from human videos or real observations into simulation. HUMAN2SIM2ROBOT~\cite{lum2025human2sim2robot}, X-SIM~\cite{dan2025xsim}, and Video2Sim2Real~\cite{han2026video2sim2real} use object motion or human demonstrations to train embodiment-specific policies, but mainly target fixed-base manipulation and require additional scene or object scans. DemoGen~\cite{xue2025demogen} generates synthetic visuomotor demonstrations, while SIMPACT~\cite{liu2026simpact}, RoboSnap~\cite{zhang2026robosnap}, and V-Dreamer~\cite{he2026vdreamer} use reconstructed or generated simulations for planning, data generation, or evaluation. Most of these methods treat the manipulated objects as rigid bodies or recover generic scene geometry, without using articulated kinematics as an explicit task representation. Moreover, simulation is typically used for only one stage of the pipeline, rather than jointly supporting instance-specific reconstruction, executable demonstration generation, and closed-loop mobile deployment.

As summarized in Table~\ref{tab:system_comparison}, prior work has separately investigated real-to-sim-to-real manipulation, articulated door opening, and legged door traversal. In contrast, our framework uses door articulation as a common task representation, connecting a digital twin reconstructed from a single RGB video with simulator-verified expert generation and onboard depth-based policy execution for complete doorway traversal.

\begin{figure*}[t]
    \vspace{2mm}
    \centering
    \includegraphics[width=\textwidth]{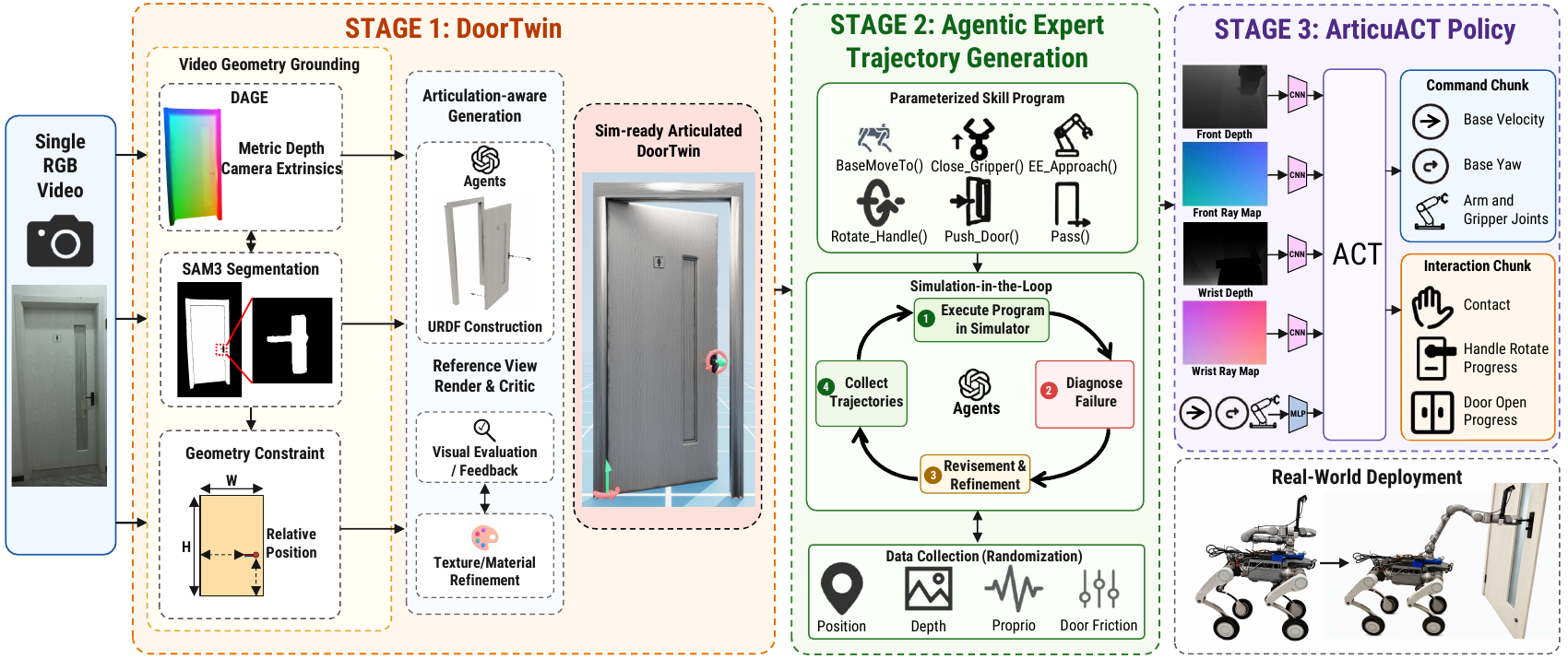}
    \caption{Overview of Video2DoorTraversal. Given a single RGB video, DoorTwin reconstructs a metrically grounded, articulated, and simulation-ready door asset through geometric grounding, articulation-aware generation, reference-view critique, and appearance refinement. A simulation-in-the-loop agent then instantiates and refines a parameterized skill program to collect randomized expert trajectories. Finally, ArticuACT combines dual-view depth observations with robot-centric Pl\"ucker ray maps to predict coordinated base, arm, and gripper commands for real-world door opening and traversal.}
    \label{fig:structure-overview}
\end{figure*}

\section{METHOD}

As illustrated in Fig.~\ref{fig:structure-overview}, our framework transforms a single RGB video of a real door into a deployable quadruped loco-manipulation policy through three stages. First, \textbf{DoorTwin} reconstructs a metrically aligned, articulated, and simulation-ready door asset. Second, a simulation-in-the-loop agent synthesizes and iteratively refines parameterized skill programs, from which successful domain-randomized expert trajectories are collected. Finally, \textbf{ArticuACT} learns whole-body door traversal from depth observations, using robot-centric Pl\"ucker conditioning and auxiliary interaction-state supervision. During deployment, the policy relies only on onboard depth observations and proprioceptive states, without privileged information.

\subsection{DoorTwin: One-shot Sim-Ready Door Generation}

Given a single RGB video of a real door, DoorTwin generates a metrically aligned, articulated, and simulation-ready door twin that preserves the geometry and appearance of the observed instance.  We first recover metric scene geometry and camera motion from the video, and then use the recovered geometry to ground articulated asset generation. The output contains an articulated door assembly, collision geometry, and textured visual meshes that can be directly imported into the physics simulator.

DoorTwin builds upon Articraft~\cite{articraft}, which generates articulated assets by iteratively writing and validating object-construction programs. Although Articraft can produce semantically plausible and kinematically valid assets, conditioning it directly on RGB appearance does not ensure instance-level consistency. In particular, absolute metric scale cannot be reliably inferred from RGB appearance alone, the relative placement of small components may be inaccurate, and large differences in part scale, such as those between a door panel and a handle, can lead to poor geometry. Moreover, procedural materials are often substantially simpler than the observed appearance. DoorTwin addresses these limitations through video geometry grounding, articulation-aware program generation, reference view rendering and critique, and appearance transfer.

\textbf{Video Geometry Grounding:}
We first process the input video using DAGE~\cite{ngo2026dage}, which estimates metrically scaled geometry and camera poses from uncalibrated RGB video. This provides a metric depth for each frame together with the corresponding camera extrinsics. For doors containing glass or reflective regions, where the DAGE depth prediction can become unreliable, we further refine the estimated depth
using LingBot-Depth~\cite{tan2026masked}.

We then apply SAM~3~\cite{carion2025sam3} to every video frame to segment the complete door assembly. The masked metric geometry from each frame is transformed into a common reference frame using the estimated camera poses:
\begin{equation}
\mathbf{x}_{i,u}
=
\mathbf{T}_{i}\mathbf{p}_{i,u},
\qquad
u \in M_i ,
\label{eq:video_door_geometry}
\end{equation}
where $\mathbf{p}_{i,u}$ denotes the metric 3D point associated with pixel $u$ in frame $i$, $\mathbf{T}_{i}$ transforms the camera coordinate system to the common door frame, and $M_i$ is the corresponding door mask. Aggregating the masked geometry across frames produces a metrically scaled multi-view point cloud of the complete door.

We establish a global coordinate frame from the aggregated door point cloud. Principal component analysis provides three dominant directions. The shortest direction approximately corresponds to the door surface normal, while the remaining two in-plane directions define the door width and height. Compared with a single view, aggregating observations across the video provides more complete geometric evidence and enables more stable estimation of the global door dimensions.

We use a VLM to select the most proper frame with the complete door mask as the reference frame. This frame is then used as the visual input to Articraft and for subsequent rendering and critique.
On the selected reference frame, we additionally segment the door panel and handle to recover their local geometry. Let $\mathbf{c}_{p}$ and $\mathbf{c}_{h}$ denote the centers of the corresponding 3D bounding boxes, and let $R_p$ denote the panel coordinate frame. The panel-relative handle location is represented as
\begin{equation}
\Delta \mathbf{p}_{h}
=
R_p^{\top}
\left(
\mathbf{c}_{h}-\mathbf{c}_{p}
\right).
\label{eq:handle_relative_position}
\end{equation}
The aggregated door dimensions and panel-relative handle location form the primary geometric constraints supplied to the asset-generation agent.

\textbf{Articulation-Aware Program Generation:}
The recovered constraints are converted into a structured specification for the Articraft agent. The specification defines the metric dimensions of the complete door, the panel-relative handle location, and the required articulation. In particular, the frame remains fixed, the door panel is connected to the frame through a revolute joint, and the handle is attached to the panel at the grounded location.

Asset generation is performed in a coarse-to-fine manner. The agent first constructs the global door assembly under the metric constraints and subsequently refines the smaller handle component and its attachment. This ordering reduces errors caused by the substantial scale difference between the panel and the handle.

We retain the native procedural validators of Articraft to check program execution, disconnected or floating components, mesh interpenetration, and invalid joint definitions. These inherited checks ensure that the generated program is structurally valid and physically loadable, but they do not determine whether the resulting asset resembles the specific door shown in the input video.

\textbf{Reference View Rendering and Critique:}
The procedural checks inherited from Articraft ensure structural validity but do not guarantee instance-level similarity. We therefore introduce a reference view rendering and critique loop. The camera intrinsics of the reference frame are estimated from the recovered scene geometry, camera pose, and their 2D--3D correspondences. The generated asset is rendered from the camera viewpoint of the selected reference frame and compared with its masked RGB image.

A visual critic evaluates the global silhouette, handle type, hinge side, relative proportions, and handle placement. When a mismatch is detected, structured feedback is returned to the generation agent, which updates the construction program while keeping the video-derived metric constraints fixed. This process is repeated until the asset passes both procedural validation and the reference view similarity check.

\textbf{Delighting and Appearance Transfer:}
After the articulated geometry has been finalized, we transfer the appearance of the observed door without modifying its validated geometry or joints. Since the selected RGB frame contains illumination and shadows from the capture environment, we use a GPT-based image editing procedure to remove these lighting effects and obtain a cleaner appearance reference. The finalized untextured door model and the processed reference image are then provided to Tripo 3D~\cite{tripo3d} to synthesize the texture and material maps.

The resulting DoorTwin is therefore grounded by metric video geometry, validated as an executable articulated asset, explicitly matched to the observed instance through reference view critique, and visually aligned with the real door for sim-to-real policy training and evaluation.




\subsection{Agentic Expert Trajectory Generation}

Given a reconstructed door twin $\Theta_D$, we generate expert
demonstrations using a simulation-in-the-loop agent. Rather than
synthesizing low-level controller code, the agent instantiates an
interpretable, parameterized skill program
\begin{equation}
    \Pi_D =
    \left\{(\sigma_j, \eta_j)\right\}_{j=1}^{M},
    \qquad
    \sigma_j \in \mathcal{S},\;
    \eta_j \in \mathcal{H}_j,
\end{equation}
where $\mathcal{S}$ contains
\textsc{BaseMoveTo}, \textsc{EE\_Approach},
\textsc{Close\_Gripper}, \textsc{Rotate\_Handle}, \textsc{Push\_Door}, \textsc{Pass}, and
\textsc{ReleaseAndRetract}. Each $\eta_j$ specifies a compact set of
task-level parameters, including the approach distance, grasp offset,
handle rotation, contact bias, base velocity, and phase duration.
Interaction targets are represented in a handle-local
approach--lateral--vertical frame, while the hinge orientation
determines the valid pushing direction.

The program is executed in parallel Isaac Gym environments at
$50\,\mathrm{Hz}$:
\begin{equation}
    x_{t+1}
    =
    \operatorname{Sim}
    \left(x_t, a_t; \Theta_D, \Theta_R\right).
\end{equation}
Each rollout records task-progress signals, including handle rotation,
door angle, unlocking state, and body passage, together with feasibility
signals. Upon failure, the agent receives a structured
diagnostic summary and selected multi-view keyframes, attributes the
failure to its most likely cause, and proposes bounded modifications to
$\eta_j$. A local simulator search subsequently evaluates neighboring
candidates, and accepts only parameterizations satisfying the task,
collision, and kinematic constraints. This
generate--execute--diagnose--refine loop continues until a successful
rollout is obtained or the search budget is exhausted.

Only successful rollouts are retained as demonstrations:
\begin{equation}
    \tau =
    \left\{
    D_t^{f}, D_t^{w}, s_t, a_t^{\star}, \phi_t, z_t
    \right\}_{t=1}^{T},
\end{equation}
where $a_t^{\star}$ is the expert action, $\phi_t$ denotes the active
skill phase, and $z_t$ contains simulator-only annotations used for
trajectory filtering and evaluation. The trajectories are recorded at
$25\,\mathrm{Hz}$ for policy learning. 

\textbf{Data Collection:}
After validating an expert program, we replay it under randomized robot, scene, and sensor configurations to collect diverse demonstrations while preserving the nominal geometry of the reconstructed door twin. We randomize the initial base and door poses, hinge and handle friction and damping, door-opening resistance, and camera extrinsics. Simulated and real depth maps are clipped to $[0.2,1.5]$~m; during training, they are further augmented with distance-dependent Gaussian noise, edge corruption, blockwise holes, pixel dropout, salt-and-pepper noise, and Gaussian blur. The front camera captures global context for door approach and free-space estimation, while the wrist camera provides local handle geometry under self-occlusion. Only rollouts that remain successful under these perturbations are retained for robust sim-to-real policy learning.

\subsection{ArticuACT Policy}
\label{sec:articuact}

We build our imitation learning policy, termed \textbf{ArticuACT}, upon the Action Chunking Transformer (ACT)~\cite{zhao2023act}. At time step $t$, the policy takes the front-view depth image $D_t^{f}$, wrist-view depth image $D_t^{w}$, and the $9$-D robot state $\mathbf{s}_t$, and predicts an action chunk of length $H=100$:
\begin{equation}
\hat{\mathbf{A}}_t
=
\pi_{\theta}
\left(
D_t^{f},
D_t^{w},
\mathbf{s}_t
\right)
\in
\mathbb{R}^{H \times 9}.
\end{equation}
Each action consists of the base forward velocity, base yaw velocity, six arm joint commands, and one gripper command. Compared with the original ACT, ArticuACT introduces robot-centric Pl\"ucker conditioning and future interaction-state prediction.

\textbf{Robot-Centric Pl\"ucker Conditioning:}
Following Jiang et al.~\cite{jiang2025knowyourcamera}, who explicitly
condition visuomotor policies on camera geometry using pixel-aligned
Pl\"ucker ray maps, we adapt this representation to our dual-depth
mobile-manipulation setting. In particular, we express the rays from
both the front and wrist cameras in a shared robot-base coordinate
frame, providing the policy with an explicit geometric correspondence
between image pixels and the action frame.

For each image pixel \((u,v)\), we construct a Pl\"ucker ray using the
camera intrinsic matrix \(K\) and the camera pose
\((R_{bc}, \mathbf{t}_{bc})\) with respect to the robot base frame:
\begin{equation}
\mathbf{d}_{b}
=
R_{bc}
\frac{
K^{-1}[u,v,1]^{\top}
}{
\left\|K^{-1}[u,v,1]^{\top}\right\|_{2}
},
\qquad
\mathbf{m}_{b}
=
\mathbf{t}_{bc}\times\mathbf{d}_{b}.
\label{eq:plucker_ray}
\end{equation}
The resulting per-pixel Pl\"ucker representation is
\begin{equation}
\mathbf{r}(u,v)
=
[\mathbf{d}_{b},\mathbf{m}_{b}]
\in \mathbb{R}^{6}.
\label{eq:plucker_map}
\end{equation}

Separate Pl\"ucker maps are constructed for the front and wrist cameras
using their respective intrinsics and extrinsics. Following the
late-fusion design for pretrained visual encoders~\cite{jiang2025knowyourcamera},
each map is processed by a lightweight convolutional encoder and
concatenated with the corresponding ResNet-18 depth feature before
projection to the Transformer dimension. This conditioning provides
explicit pixel-aligned camera geometry in the robot base frame and
reduces spatial ambiguity under camera-viewpoint.

\textbf{Future Interaction-State Prediction:}
In addition to robot actions, each future decoder token predicts a three-dimensional interaction state:
\begin{equation}
\hat{\mathbf{z}}_{t+h}
=
\left[
\hat{c}_{t+h},
\hat{p}^{\mathrm{handle}}_{t+h},
\hat{p}^{\mathrm{door}}_{t+h}
\right],
\end{equation}
where the three components represent stable handle contact, handle-rotation progress, and door-opening progress, respectively. ArticuACT therefore predicts an $H\times9$ action chunk together with a temporally aligned $H\times3$ interaction chunk.

The interaction predictions are used only as auxiliary supervision. They are not fed back into the action decoder and do not modify the $9$-D robot control interface. Instead, the auxiliary task encourages the shared decoder features to encode contact and task-progress information.

The complete training objective is
\begin{equation}
\mathcal{L}
=
\mathcal{L}_{\mathrm{act}}
+
\beta \mathcal{L}_{\mathrm{KL}}
+
\lambda_c \mathcal{L}_{\mathrm{contact}}
+
\lambda_h \mathcal{L}_{\mathrm{handle}}
+
\lambda_d \mathcal{L}_{\mathrm{door}},
\end{equation}
where $\mathcal{L}_{\mathrm{act}}$ is a masked $\ell_1$ action loss, $\mathcal{L}_{\mathrm{contact}}$ is a binary cross-entropy loss, and the handle and door progress terms use Smooth $\ell_1$ losses. 
Interaction losses are backpropagated via shared decoder and policy backbone, with interaction labels required only in training.

\section{EXPERIMENTS}

\subsection{Experimental Setup}

For door asset generation, we capture a single RGB video of each target door using a smartphone camera. The hardware robot platform consists of a Unitree A2-W wheel-legged base equipped with a Unitree Z1 arm. Two Intel RealSense D435 cameras provide onboard perception: one is mounted on the robot head for global scene observation, while the other is mounted near the gripper for close-range interaction sensing.
The onboard computer of the A2-W executes the low-level base controller. An additional NVIDIA Jetson Orin NX processes the visual observations, performs policy inference, and control the arm.

\begin{figure}[t]
    \vspace{2mm}
    \centering
    \includegraphics[width=\linewidth]{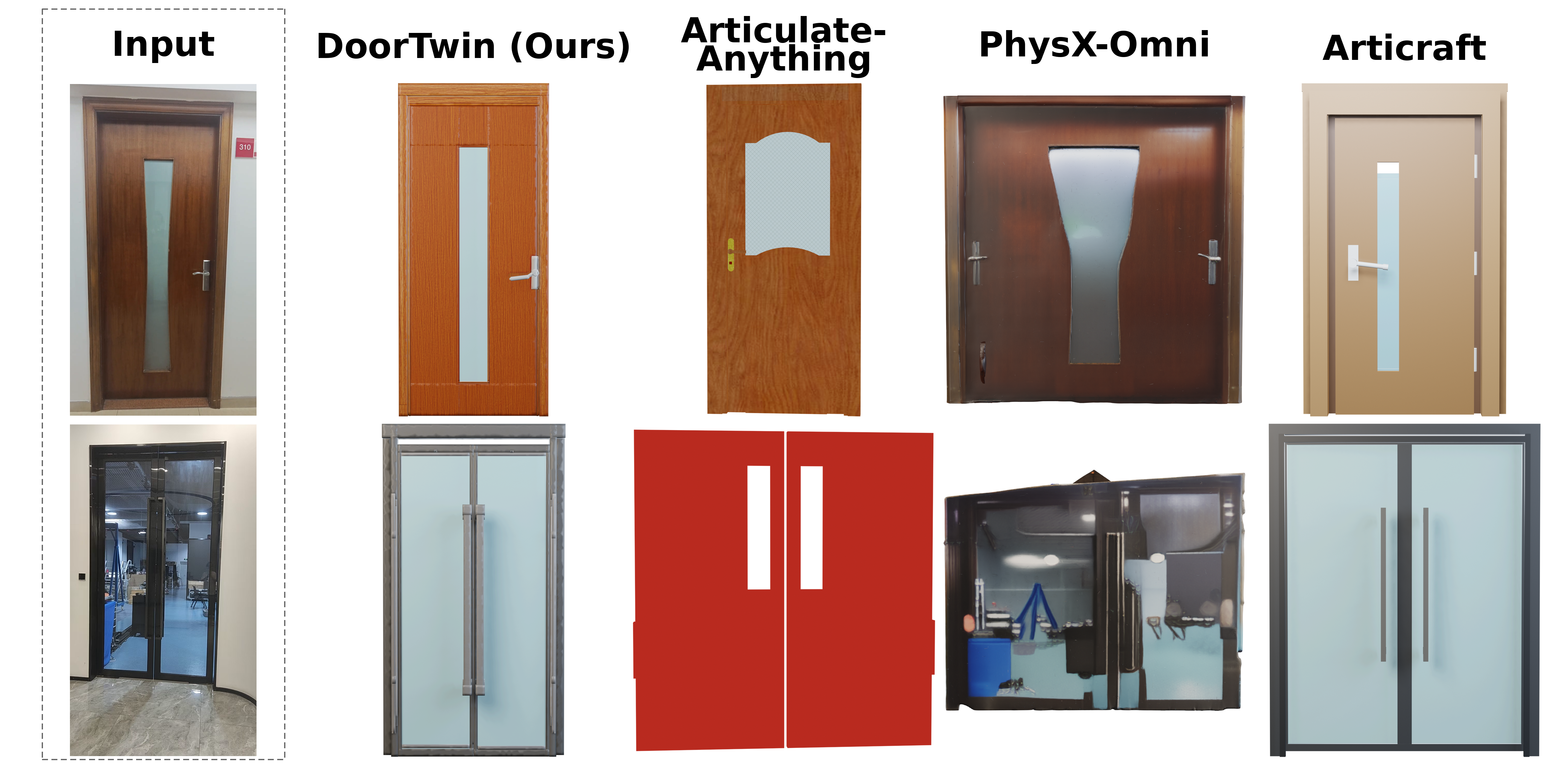}
    \caption{Qualitative comparison of articulated door assets.}
    \label{fig:assets_compare}
    \vspace{-0.2cm}
\end{figure}

\begin{figure}[t]
    \centering
    \includegraphics[width=\linewidth]{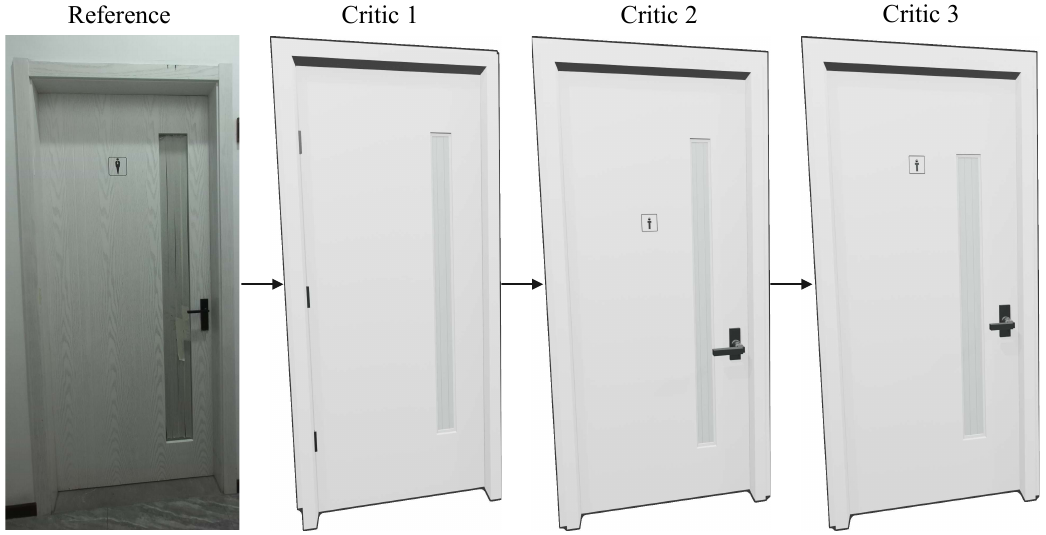}
    \caption{Reference-view rendering and critic refinement iterations.}
    \label{fig:critic}
    \vspace{-0.2cm}
\end{figure}

\subsection{Asset Generation}

We primarily evaluate whether the geometry of the generated door is consistent with that in the input image, including the global metric dimensions of the door and the panel-relative placement of the handle. We assess DoorTwin against three representative articulated-asset generation baselines: Articraft~\cite{articraft},
Articulate-Anything~\cite{articulateanything}, and
PhysX-Omni~\cite{physxomni}. For a controlled comparison,
DoorTwin, Articraft, and Articulate-Anything are evaluated using
GPT-5.5, while PhysX-Omni retains its original fine-tuned
Qwen2.5-VL model~\cite{qwen}. We conduct the asset-generation evaluation on 20 real-world door instances.

We evaluate asset quality using both geometry and RGB metrics.
For geometry, we introduce the Scale Score (SS), which measures metric
accuracy as
\begin{equation}
\mathrm{SS}
=
100\max\left(
0,\,
1-\frac{e_w+e_h+e_p}{3}
\right),
\end{equation}
where $e_w$, $e_h$, and $e_p$ are the normalized errors of door width,
door height, and handle position, respectively. Masked IoU (mIoU) measures the overlap between the rendered asset mask
and the reference door mask. For appearance, RGB PSNR, SSIM, and LPIPS are computed within the reference foreground mask. We additionally report a VLM Score, where a vision-language model evaluates the overall instance-level consistency
between the generated asset and the reference image.

\textbf{Asset Generation Results}
As shown in Table~\ref{tab:asset_generation_quality} and
Fig.~\ref{fig:assets_compare}, DoorTwin achieves the best overall
performance, with the highest Scale Score, mIoU, PSNR, and VLM
Score, as well as the lowest LPIPS. PhysX-Omni occasionally misidentifies
doors as other articulated objects, leading to substantial errors in global
geometry and part layout. Articulate-Anything generally retrieves plausible
door assets, but often exhibits mismatched panel proportions and handle
placement. Articraft produces structurally valid articulated assets, yet
still shows noticeable errors in metric scale, local components, and
appearance. The slightly lower SSIM is likely due to local appearance differences between the rendered and reference images, to which SSIM is sensitive. Overall, these results demonstrate that DoorTwin better preserves both the metric geometry and instance-specific characteristics
of the observed door.

\textbf{Effect of the Reference-View Critic}
As illustrated in Fig.~\ref{fig:critic}, the reference-view critic progressively improves the instance-level consistency of the generated asset. In the first refinement iteration, the critic recovers missing components, including the door marking and handle. The second and third iterations further correct their size, local geometry, and placement to better match the reference image. This iterative process complements procedural validation, which ensures structural validity and simulation readiness but cannot guarantee visual similarity to the observed door instance.

\begin{table}[t]
\vspace{2mm}
    \centering
    \caption{Comparison of Geometry and Appearance Quality.}
    \label{tab:asset_generation_quality}
    \resizebox{\columnwidth}{!}{
    \begin{tabular}{l|cc|ccc|c}
        \toprule
        \multirow[c]{2}{*}[-1.0ex]{\textbf{Method}}
        & \multicolumn{2}{c|}{\textbf{Geometry}}
        & \multicolumn{3}{c|}{\textbf{Appearance}}
        & \multirow[c]{2}{*}[-1.0ex]{\textbf{VLM Score $\uparrow$}} \\
        
        \cmidrule(lr){2-3}
        \cmidrule(lr){4-6}
        
        & \textbf{SS $\uparrow$}
        & \textbf{mIoU $\uparrow$}
        & \textbf{PSNR $\uparrow$}
        & \textbf{SSIM $\uparrow$}
        & \textbf{LPIPS $\downarrow$}
        & \\
        \midrule

        PhysX-Omni
        & 65.75 & 0.635 & 17.51 & \textbf{0.726} & 0.484 & 11.54 \\

        Articraft
        & 89.03 & 0.880 & 16.64 & 0.702 & 0.470 & 28.75 \\

        Articulate Anything
        & 86.62 & 0.831 & 16.03 & 0.667 & 0.565 & 11.59 \\

        \textbf{DoorTwin (Ours)}
        & \textbf{94.95}
        & \textbf{0.972}
        & \textbf{18.53}
        & 0.692
        & \textbf{0.408}
        & \textbf{56.74} \\

        \bottomrule
    \end{tabular}
    }
\end{table}

\subsection{Simulation Results}

\textbf{Baselines:}
We compare our method with the following baselines:
1) \textbf{Replay}, which open-loop replays one generated trajectory by directly setting the recorded actions as control targets;
2) \textbf{Object-Aware (OA) Replay}~\cite{lum2025human2sim2robot}, which uses SAM~3~\cite{carion2025sam3} to segment the door handle and FoundationPose~\cite{wen2024foundationpose} to estimate its 6-DoF pose, and then warps the reference trajectory according to the relative transformation between the nominal and estimated handle poses;
3) \textbf{DoorGym}~\cite{urakami2019doorgym}, for which we adapt its state-based PPO implementation as a reinforcement-learning baseline. Following its original setting, a floating gripper is initialized near the handle and trained using privileged simulator states;
4) \textbf{UniDoorManip}~\cite{li2024unidoormanip}, which uses third-person point-cloud observations and a hierarchical policy for door manipulation;
5) \textbf{Vanilla ACT}~\cite{zhao2023act}, implemented with the original action-chunking architecture and our dual-depth observation and action interfaces;
6) \textbf{Diffusion Policy (DP)}~\cite{chi2023diffusionpolicy}, a widely used visuomotor policy that models action sequences through conditional denoising diffusion;
7) \textbf{DP3}~\cite{ze2024dp3}, which encodes point-cloud observations and predicts a sequence of delta end-effector transformations.

\textbf{Implementation details:}
For each door instance, we collect 200 successful demonstrations for policy training. We evaluate all methods on 20 door instances using 256 trials with independently sampled random seeds. The evaluation uses the same randomization ranges as training. All learning-based policies predict high-level commands at 25 Hz.

\textbf{Evaluation Metrics:}
We report the door-opening success rate and the full-traversal success rate. A trial is considered a successful door opening only if the handle rotation first reaches the predefined unlocking threshold and the door angle subsequently exceeds $80^\circ$ within the $20\,\mathrm{s}$ time limit. Full traversal additionally requires the robot base to move at least $1\,\mathrm{m}$ beyond the doorway; therefore, traversal success is counted only after successful door opening. A timeout or any collision between the robot base and the door before unlocking is considered a failure.

\textbf{Results:}
As shown in Table~\ref{tab:simulation_results}, our method achieves the highest success rates for both door opening and traversal. OA Replay substantially improves over direct trajectory replay, but remains sensitive to handle-pose estimation errors; even a small localization error can shift the entire transferred trajectory and lead to inaccurate grasping. DoorGym performs poorly because the long-horizon task involves a large multi-stage exploration space, while designing balanced rewards for grasping, unlocking, pushing, and traversal is difficult. UniDoorManip achieves reasonable door-opening performance but degrades significantly during traversal, which may be related to its third-person point-cloud observation being less effective for the precise base--arm coordination required after opening the door. Among imitation-learning baselines, our method consistently outperforms vanilla ACT, Diffusion Policy, and DP3, demonstrating the effectiveness of the proposed geometry- and interaction-aware policy design.

\begin{table}[t]
\vspace{2mm}
    \centering
    \caption{Simulation Results on the Door-Traversal Task.}
    \label{tab:simulation_results}
    \resizebox{\columnwidth}{!}{
    \begin{tabular}{l|c|c}
        \toprule
        \textbf{Method}
        & \textbf{Door Opening Success $\uparrow$}
        & \textbf{Traversal Success $\uparrow$} \\
        \midrule
        Replay
        & 14.06\%
        & 14.06\% \\

        OA Replay
        & 53.13\%
        & 53.13\% \\

        DoorGym
        & 12.50\%
        & 12.50\% \\

        UniDoorManip
        & 74.22\%
        & 50.78\% \\
        \midrule
        Vanilla ACT
        & 67.58\%
        & 64.84\% \\

        Diffusion Policy (DP)
        & 64.45\%
        & 64.45\% \\

        DP3
        & 68.36\%
        & 66.41\% \\

        \textbf{Ours}
        & \textbf{98.44\%}
        & \textbf{97.27\%} \\
        \bottomrule
    \end{tabular}
    }
\end{table}

\subsection{Ablation Study}

\textbf{Policy Components and Command-Space Ablation:}
We evaluate the contributions of Pl\"ucker conditioning and interaction-state prediction under both end-effector and joint-command action spaces. Specifically, we compare vanilla ACT, ACT with either individual module, and the complete ArticuACT policy.

\begin{figure}[t]
    \vspace{2mm}
    \centering
    \begin{minipage}[c]{0.49\linewidth}
        \centering
        \includegraphics[width=\linewidth]
        {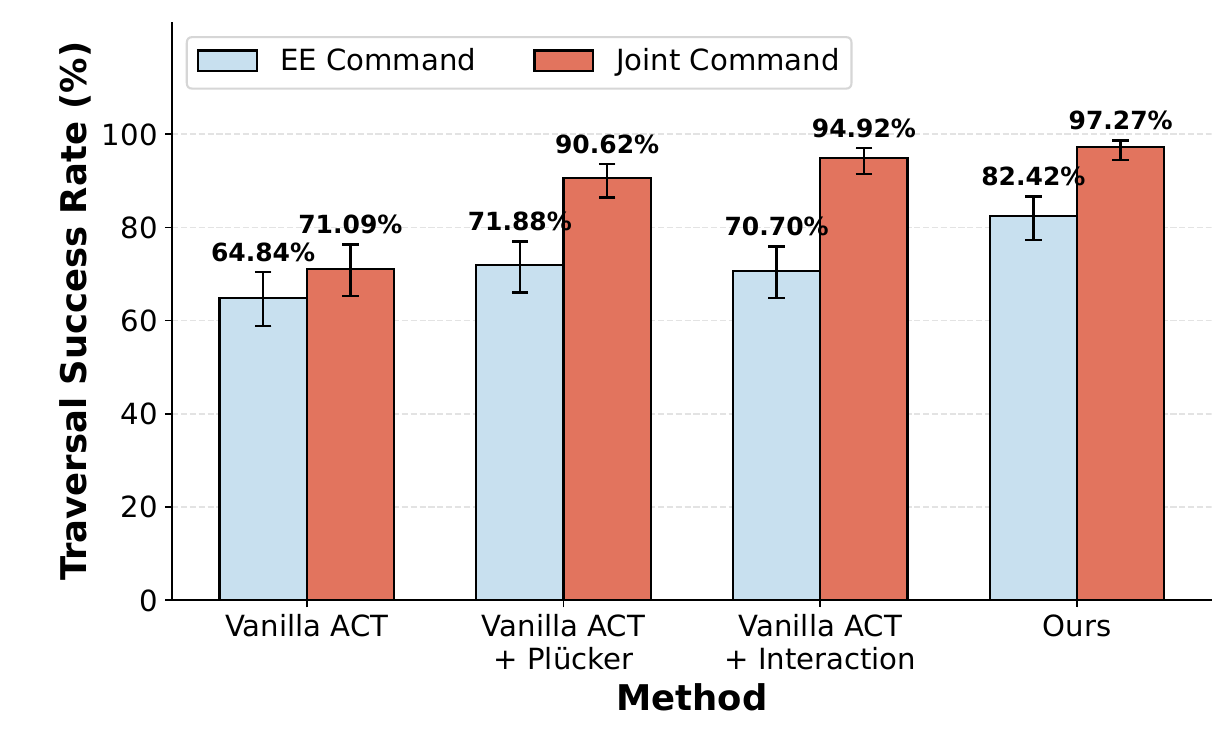}
    \end{minipage}
    \hfill
    \begin{minipage}[c]{0.49\linewidth}
        \centering
        \includegraphics[width=\linewidth]
        {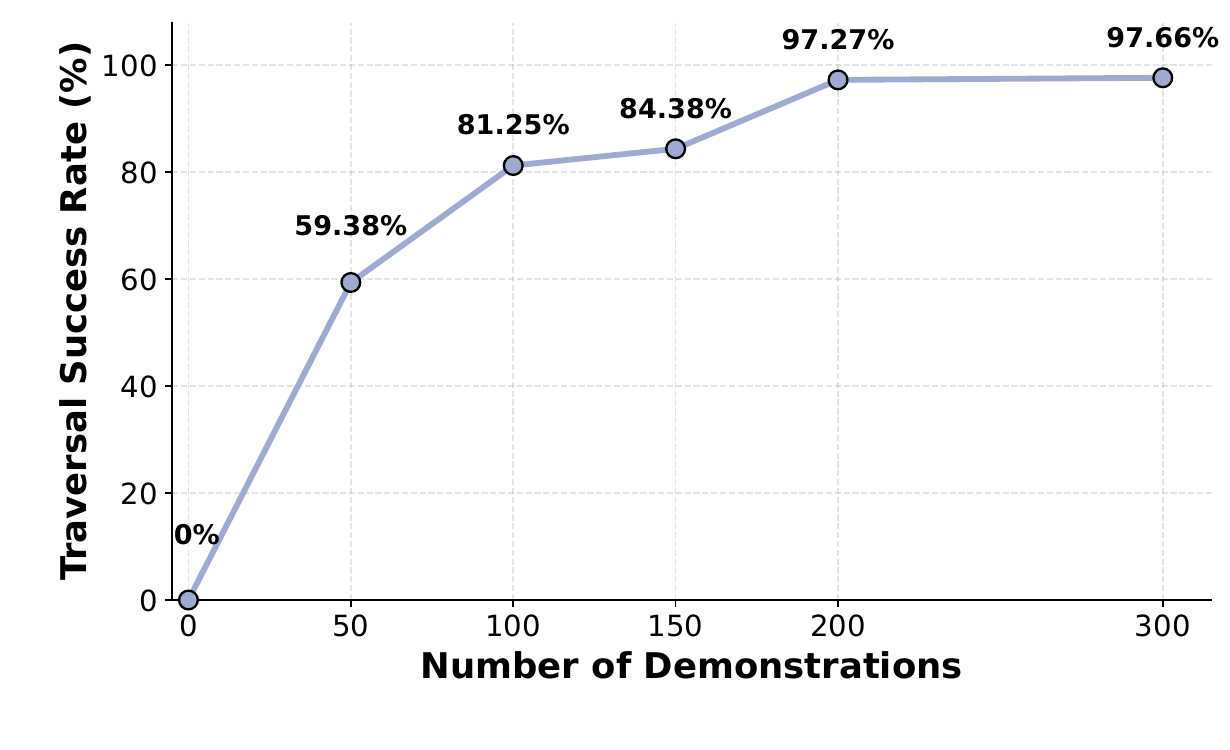}
    \end{minipage}
    \caption{Left: ablation study of the proposed policy components and command spaces. Right: scaling performance of our method.}
    \label{fig:policy_ablation_and_scaling}
\end{figure}

As shown in Fig.~\ref{fig:policy_ablation_and_scaling}, both proposed modules consistently improve traversal performance over vanilla ACT, while their combination achieves the best results in both command spaces. Compared with vanilla ACT, the complete model improves the traversal success rate by $17.58\%$ with end-effector commands and $26.18\%$ with joint commands. It further outperforms the strongest single-module variant by $2.35\%$ under joint-command control, demonstrating that robot-centric geometric conditioning and interaction-aware auxiliary supervision provide complementary benefits. Moreover, joint-command policies consistently outperform their end-effector-command counterparts, suggesting that direct joint prediction is better aligned with the low-level controller and enables more reliable execution during precise and contact-rich door traversal.

\textbf{Agentic Trajectory Generation Ablation:}
We evaluate the agentic trajectory generation framework against three variants. The rule-based baseline adapts a fixed reference trajectory by translating the entire trajectory according to the displacement between the nominal and detected handle centers, while keeping its motion pattern unchanged. We further compare our method without simulator rollout feedback and a variant that uses rollout logs for VLM-based refinement.

As shown in Table~\ref{tab:agentic_ablation}, removing simulator feedback causes a substantial performance drop, showing that geometric alignment or VLM reasoning alone cannot reliably produce physically executable trajectories. The full method improves the success rate by $11.25\%$ over the rule-based baseline and by $37.50\%$ over the no-rollout variant. It also achieves the highest success rate with fewer refinement loops than log-based repair, demonstrating that simulator verification and iterative agent refinement jointly enable more reliable and efficient expert trajectory generation.

\begin{table}[t]
    \centering
    \caption{Ablation Study of the Agentic Trajectory Generation.}
    \label{tab:agentic_ablation}
    \resizebox{\columnwidth}{!}{
    \begin{tabular}{l|c|c}
        \toprule
        \textbf{Method}
        & \textbf{Success Rate $\uparrow$}
        & \textbf{Avg. Agent Loops $\downarrow$} \\
        \midrule

        Rule-based
        & 74.38\%
        & 0 \\

        Ours w/o Simulation Rollout
        & 48.13\%
        & 0 \\

        Ours w/ VLM Fix--Log
        & 85.00\%
        & 2.8 \\

        \textbf{Ours Full}
        & \textbf{85.63\%}
        & 2.5 \\

        \bottomrule
    \end{tabular}
    }
\end{table}

\textbf{Data Scaling Analysis:}
As shown in Fig.~\ref{fig:policy_ablation_and_scaling}, traversal success consistently improves with more demonstrations, rising from $59.38\%$ with 50 trajectories to $97.27\%$ with 200. The largest gain occurs within the first 100 demonstrations, while additional data continues to improve policy robustness.

\begin{figure}[t]
    \vspace{2mm}
    \centering
    \includegraphics[width=\linewidth]{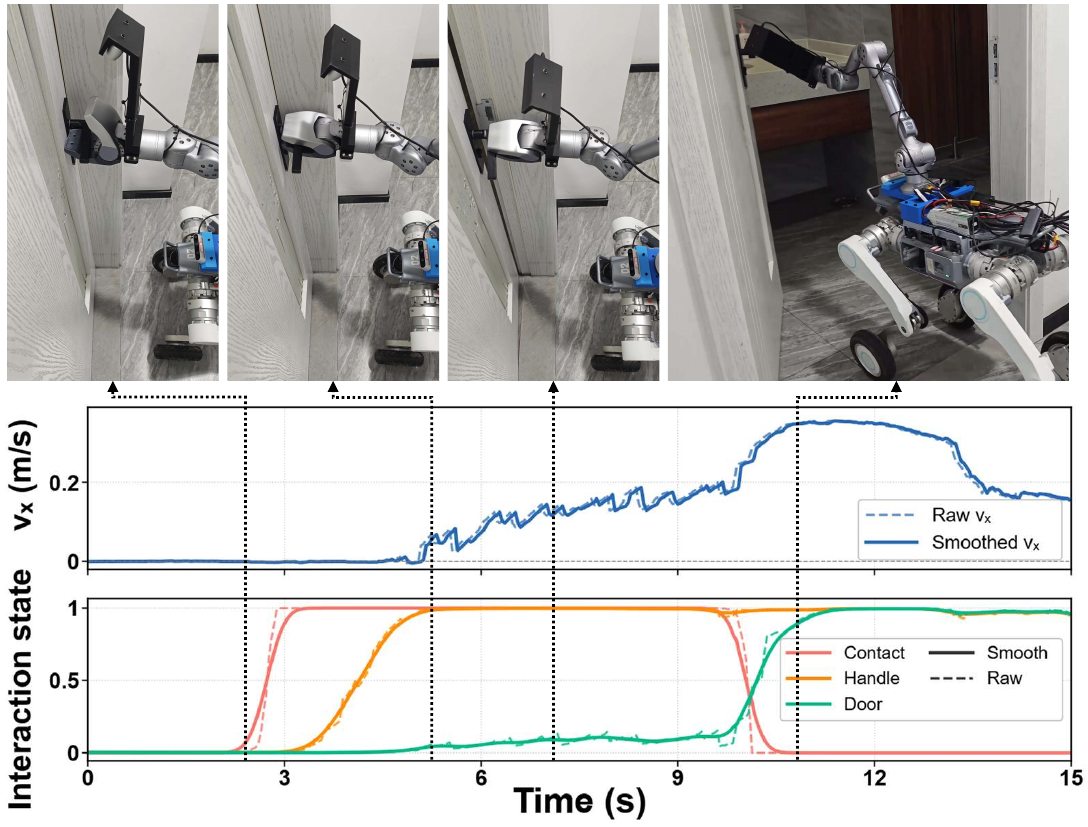}
    \caption{Snapshots of real-world experiments with interaction state.}
    \label{fig:real}
\end{figure}

\begin{table}[t]
    \centering
    \caption{Real-World Results Across Five Doors.}
    \label{tab:cross_door_results}
    \resizebox{\columnwidth}{!}{
    \begin{tabular}{l|c|c|c|c|c|c}
        \toprule
        \textbf{Method}
        & \textbf{Door 1}
        & \textbf{Door 2}
        & \textbf{Door 3}
        & \textbf{Door 4}
        & \textbf{Door 5}
        & \textbf{Average Success Rate} \\
        \midrule

        Replay
        & 6/35
        & 5/35
        & 5/35
        & 8/35
        & 8/35
        & 18.29\% \\

        OA Replay
        & 10/35
        & 12/35
        & 9/35
        & 15/35
        & 18/35
        & 36.58\% \\

        \midrule

        Vanilla ACT
        & 25/35
        & 20/35
        & 23/35
        & 25/35
        & 22/35
        & 65.71\% \\

        \textbf{Ours}
        & \textbf{35/35}
        & \textbf{32/35}
        & \textbf{35/35}
        & \textbf{33/35}
        & \textbf{34/35}
        & \textbf{96.57\%} \\

        \bottomrule
    \end{tabular}
    }
    \vspace{-0.2cm}
\end{table}

\subsection{Real-World Results}

As shown in Table~\ref{tab:cross_door_results}, our method achieves consistent success across all five door instances and substantially outperforms the replay-based baselines and vanilla ACT. The results demonstrate that the proposed framework generalizes reliably across different door geometries while maintaining stable performance throughout the complete opening and traversal process.

As illustrated in Fig.~\ref{fig:real}, the predicted
interaction states are temporally aligned with the physical task
progression. The contact signal activates during handle engagement,
followed by handle-rotation and door-opening progress, while the
forward base velocity increases smoothly as the robot transitions
toward traversal. This alignment indicates that ArticuACT captures
meaningful interaction phases during real-world execution.

\textbf{Execution Efficiency:}
Our mobile manipulator completes the full approach, opening, and
traversal sequence in only approximately \(13\,\mathrm{s}\) on average,
achieving an execution speed comparable to related systems. This efficiency results from the coordinated prediction of base and arm commands, allowing the robot to manipulate and traverse the door continuously without relying on slow stage transitions.

\textbf{Zero-Shot Cross-Door Generalization:}
Although our framework follows a single-video real-to-sim-to-real setting, we further evaluate whether a policy trained on one reconstructed door can transfer directly to structurally similar unseen doors without additional training or trajectory generation. Across three unseen doors, the policy succeeds in 25/35, 31/35, and 29/35 trials, respectively, yielding an average zero-shot success rate of 80.95\%. These results demonstrate that the learned policy exhibits meaningful zero-shot generalization across similar door structures.


\section{Conclusion}

We presented \textbf{Video2DoorTraversal}, a single-video real-to-sim-to-real framework for wheel-legged door opening and traversal. From one RGB video, \textbf{DoorTwin} reconstructs an instance-aligned, articulated, and simulation-ready door twin. A simulation-in-the-loop agent then generates physically executable traversal demonstrations, which are used to train \textbf{ArticuACT} for base-arm coordination from onboard depth observations. Extensive experiments validate the complete pipeline, achieving \(96.57\%\) average success across five real doors. The learned policy further achieves \(80.95\%\) zero-shot success on structurally similar unseen doors. Future work will extend the framework to pull doors, additional handle mechanisms, and more diverse door geometries.

\bibliographystyle{IEEEtran}
\bibliography{references}

\end{document}